\documentclass[letterpaper, 10 pt, conference]{ieeeconf}  

\IEEEoverridecommandlockouts                              
\usepackage[utf8]{inputenc}
\usepackage[T1]{fontenc}

\title{\LARGE \bf
AM-RRT*: Informed Sampling-based Planning with Assisting Metric
}

\usepackage{algorithm}
\usepackage{hyperref}
\usepackage[noend]{algpseudocode}
\usepackage{amssymb}

\usepackage{amsmath}
\usepackage{caption}
\usepackage{setspace}
\usepackage{pgfplots}
\usepackage[
    backend=biber,
    style=numeric,
    sorting=none
]{biblatex}
\newcommand\blfootnote[1]{%
  \begingroup
  \renewcommand\thefootnote{}\footnote{#1}%
  \addtocounter{footnote}{-1}%
  \endgroup
}
\author{Daniel Armstrong$^*$\\ \small Dept. of Computer Science\\ University of Oxford\\ \href{mailto:daniel.armstrong@cs.ox.ac.uk}{daniel.armstrong@cs.ox.ac.uk} 
    \and 
    André Jonasson\\ \small Ocado Technology 10x\\ \href{mailto:andre.jonasson@ocado.com}{andre.jonasson@ocado.com} }

\begin{document}

\twocolumn[{
\renewcommand\twocolumn[1][]{#1}
\maketitle
    \centering
    \resizebox {0.91\textwidth} {!} {
\begingroup%
\makeatletter%
\begin{pgfpicture}%
\pgfpathrectangle{\pgfpointorigin}{\pgfqpoint{12.400000in}{2.700000in}}%
\pgfusepath{use as bounding box, clip}%
\begin{pgfscope}%
\pgfsetbuttcap%
\pgfsetmiterjoin%
\definecolor{currentfill}{rgb}{1.000000,1.000000,1.000000}%
\pgfsetfillcolor{currentfill}%
\pgfsetlinewidth{0.000000pt}%
\definecolor{currentstroke}{rgb}{1.000000,1.000000,1.000000}%
\pgfsetstrokecolor{currentstroke}%
\pgfsetdash{}{0pt}%
\pgfpathmoveto{\pgfqpoint{0.000000in}{0.000000in}}%
\pgfpathlineto{\pgfqpoint{12.400000in}{0.000000in}}%
\pgfpathlineto{\pgfqpoint{12.400000in}{2.700000in}}%
\pgfpathlineto{\pgfqpoint{0.000000in}{2.700000in}}%
\pgfpathclose%
\pgfusepath{fill}%
\end{pgfscope}%
\begin{pgfscope}%
\pgfpathrectangle{\pgfqpoint{0.372222in}{0.000000in}}{\pgfqpoint{2.700000in}{2.700000in}}%
\pgfusepath{clip}%
\pgfsys@transformshift{0.372222in}{0.000000in}%
\pgftext[left,bottom]{\pgfimage[interpolate=true,width=2.700000in,height=2.700000in]{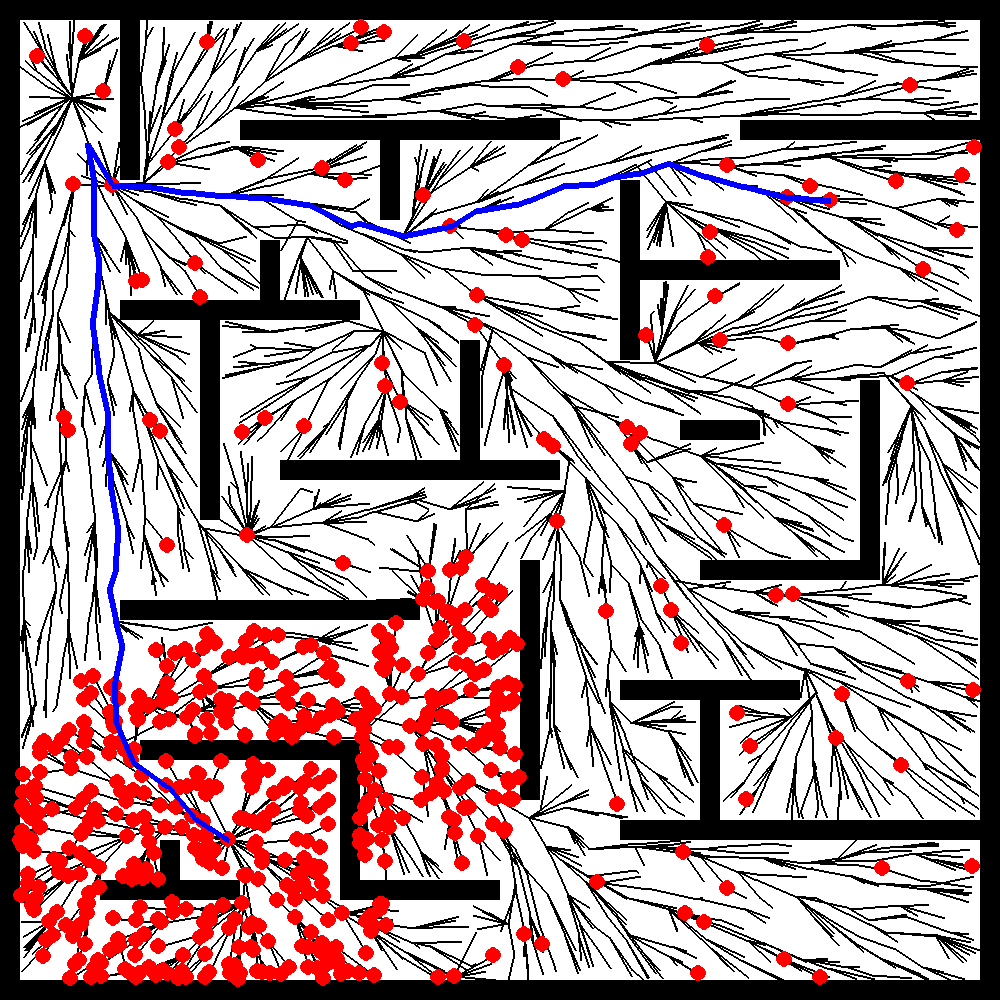}}%
\end{pgfscope}%
\begin{pgfscope}%
\pgfsetbuttcap%
\pgfsetmiterjoin%
\definecolor{currentfill}{rgb}{0.000000,0.741176,0.666667}%
\pgfsetfillcolor{currentfill}%
\pgfsetlinewidth{0.000000pt}%
\definecolor{currentstroke}{rgb}{0.000000,0.000000,0.000000}%
\pgfsetstrokecolor{currentstroke}%
\pgfsetstrokeopacity{0.000000}%
\pgfsetdash{}{0pt}%
\pgfpathmoveto{\pgfqpoint{1.004359in}{0.320934in}}%
\pgfpathlineto{\pgfqpoint{1.092786in}{0.320934in}}%
\pgfpathquadraticcurveto{\pgfqpoint{1.126119in}{0.320934in}}{\pgfqpoint{1.126119in}{0.354267in}}%
\pgfpathlineto{\pgfqpoint{1.126119in}{0.453033in}}%
\pgfpathquadraticcurveto{\pgfqpoint{1.126119in}{0.486366in}}{\pgfqpoint{1.092786in}{0.486366in}}%
\pgfpathlineto{\pgfqpoint{1.004359in}{0.486366in}}%
\pgfpathquadraticcurveto{\pgfqpoint{0.971025in}{0.486366in}}{\pgfqpoint{0.971025in}{0.453033in}}%
\pgfpathlineto{\pgfqpoint{0.971025in}{0.354267in}}%
\pgfpathquadraticcurveto{\pgfqpoint{0.971025in}{0.320934in}}{\pgfqpoint{1.004359in}{0.320934in}}%
\pgfpathclose%
\pgfusepath{fill}%
\end{pgfscope}%
\begin{pgfscope}%
\definecolor{textcolor}{rgb}{0.000000,0.000000,0.000000}%
\pgfsetstrokecolor{textcolor}%
\pgfsetfillcolor{textcolor}%
\pgftext[x=1.048572in,y=0.403650in,,]{\color{textcolor}\rmfamily\fontsize{8.000000}{9.600000}\selectfont A}%
\end{pgfscope}%
\begin{pgfscope}%
\pgfsetbuttcap%
\pgfsetmiterjoin%
\definecolor{currentfill}{rgb}{0.000000,0.741176,0.666667}%
\pgfsetfillcolor{currentfill}%
\pgfsetlinewidth{0.000000pt}%
\definecolor{currentstroke}{rgb}{0.000000,0.000000,0.000000}%
\pgfsetstrokecolor{currentstroke}%
\pgfsetstrokeopacity{0.000000}%
\pgfsetdash{}{0pt}%
\pgfpathmoveto{\pgfqpoint{2.568266in}{2.075934in}}%
\pgfpathlineto{\pgfqpoint{2.660879in}{2.075934in}}%
\pgfpathquadraticcurveto{\pgfqpoint{2.694212in}{2.075934in}}{\pgfqpoint{2.694212in}{2.109267in}}%
\pgfpathlineto{\pgfqpoint{2.694212in}{2.208033in}}%
\pgfpathquadraticcurveto{\pgfqpoint{2.694212in}{2.241366in}}{\pgfqpoint{2.660879in}{2.241366in}}%
\pgfpathlineto{\pgfqpoint{2.568266in}{2.241366in}}%
\pgfpathquadraticcurveto{\pgfqpoint{2.534932in}{2.241366in}}{\pgfqpoint{2.534932in}{2.208033in}}%
\pgfpathlineto{\pgfqpoint{2.534932in}{2.109267in}}%
\pgfpathquadraticcurveto{\pgfqpoint{2.534932in}{2.075934in}}{\pgfqpoint{2.568266in}{2.075934in}}%
\pgfpathclose%
\pgfusepath{fill}%
\end{pgfscope}%
\begin{pgfscope}%
\definecolor{textcolor}{rgb}{0.000000,0.000000,0.000000}%
\pgfsetstrokecolor{textcolor}%
\pgfsetfillcolor{textcolor}%
\pgftext[x=2.614572in,y=2.158650in,,]{\color{textcolor}\rmfamily\fontsize{8.000000}{9.600000}\selectfont G}%
\end{pgfscope}%
\begin{pgfscope}%
\pgfpathrectangle{\pgfqpoint{4.850000in}{0.000000in}}{\pgfqpoint{2.700000in}{2.700000in}}%
\pgfusepath{clip}%
\pgfsys@transformshift{4.850000in}{0.000000in}%
\pgftext[left,bottom]{\pgfimage[interpolate=true,width=2.700000in,height=2.700000in]{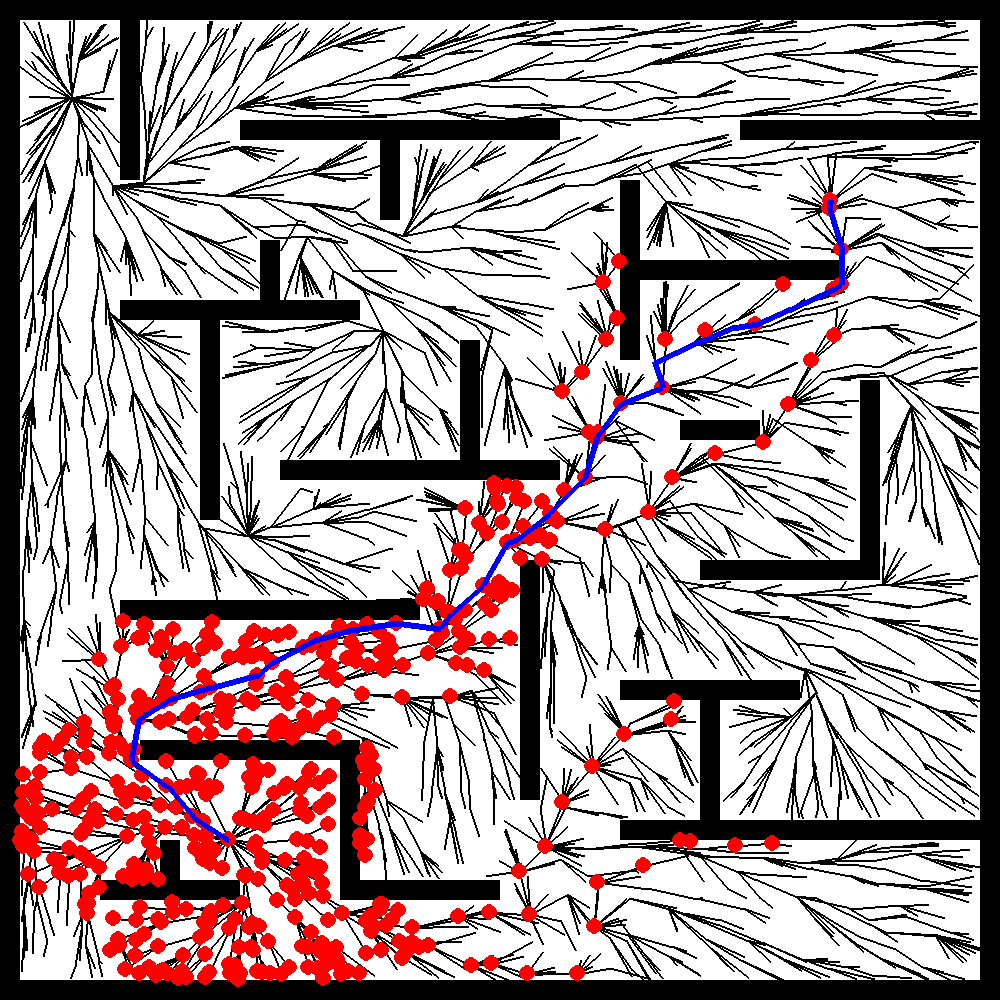}}%
\end{pgfscope}%
\begin{pgfscope}%
\pgfsetbuttcap%
\pgfsetmiterjoin%
\definecolor{currentfill}{rgb}{0.000000,0.741176,0.666667}%
\pgfsetfillcolor{currentfill}%
\pgfsetlinewidth{0.000000pt}%
\definecolor{currentstroke}{rgb}{0.000000,0.000000,0.000000}%
\pgfsetstrokecolor{currentstroke}%
\pgfsetstrokeopacity{0.000000}%
\pgfsetdash{}{0pt}%
\pgfpathmoveto{\pgfqpoint{5.482136in}{0.320934in}}%
\pgfpathlineto{\pgfqpoint{5.570564in}{0.320934in}}%
\pgfpathquadraticcurveto{\pgfqpoint{5.603897in}{0.320934in}}{\pgfqpoint{5.603897in}{0.354267in}}%
\pgfpathlineto{\pgfqpoint{5.603897in}{0.453033in}}%
\pgfpathquadraticcurveto{\pgfqpoint{5.603897in}{0.486366in}}{\pgfqpoint{5.570564in}{0.486366in}}%
\pgfpathlineto{\pgfqpoint{5.482136in}{0.486366in}}%
\pgfpathquadraticcurveto{\pgfqpoint{5.448803in}{0.486366in}}{\pgfqpoint{5.448803in}{0.453033in}}%
\pgfpathlineto{\pgfqpoint{5.448803in}{0.354267in}}%
\pgfpathquadraticcurveto{\pgfqpoint{5.448803in}{0.320934in}}{\pgfqpoint{5.482136in}{0.320934in}}%
\pgfpathclose%
\pgfusepath{fill}%
\end{pgfscope}%
\begin{pgfscope}%
\definecolor{textcolor}{rgb}{0.000000,0.000000,0.000000}%
\pgfsetstrokecolor{textcolor}%
\pgfsetfillcolor{textcolor}%
\pgftext[x=5.526350in,y=0.403650in,,]{\color{textcolor}\rmfamily\fontsize{8.000000}{9.600000}\selectfont A}%
\end{pgfscope}%
\begin{pgfscope}%
\pgfsetbuttcap%
\pgfsetmiterjoin%
\definecolor{currentfill}{rgb}{0.000000,0.741176,0.666667}%
\pgfsetfillcolor{currentfill}%
\pgfsetlinewidth{0.000000pt}%
\definecolor{currentstroke}{rgb}{0.000000,0.000000,0.000000}%
\pgfsetstrokecolor{currentstroke}%
\pgfsetstrokeopacity{0.000000}%
\pgfsetdash{}{0pt}%
\pgfpathmoveto{\pgfqpoint{7.046043in}{2.075934in}}%
\pgfpathlineto{\pgfqpoint{7.138657in}{2.075934in}}%
\pgfpathquadraticcurveto{\pgfqpoint{7.171990in}{2.075934in}}{\pgfqpoint{7.171990in}{2.109267in}}%
\pgfpathlineto{\pgfqpoint{7.171990in}{2.208033in}}%
\pgfpathquadraticcurveto{\pgfqpoint{7.171990in}{2.241366in}}{\pgfqpoint{7.138657in}{2.241366in}}%
\pgfpathlineto{\pgfqpoint{7.046043in}{2.241366in}}%
\pgfpathquadraticcurveto{\pgfqpoint{7.012710in}{2.241366in}}{\pgfqpoint{7.012710in}{2.208033in}}%
\pgfpathlineto{\pgfqpoint{7.012710in}{2.109267in}}%
\pgfpathquadraticcurveto{\pgfqpoint{7.012710in}{2.075934in}}{\pgfqpoint{7.046043in}{2.075934in}}%
\pgfpathclose%
\pgfusepath{fill}%
\end{pgfscope}%
\begin{pgfscope}%
\definecolor{textcolor}{rgb}{0.000000,0.000000,0.000000}%
\pgfsetstrokecolor{textcolor}%
\pgfsetfillcolor{textcolor}%
\pgftext[x=7.092350in,y=2.158650in,,]{\color{textcolor}\rmfamily\fontsize{8.000000}{9.600000}\selectfont G}%
\end{pgfscope}%
\begin{pgfscope}%
\pgfpathrectangle{\pgfqpoint{9.327778in}{0.000000in}}{\pgfqpoint{2.700000in}{2.700000in}}%
\pgfusepath{clip}%
\pgfsys@transformshift{9.327778in}{0.000000in}%
\pgftext[left,bottom]{\pgfimage[interpolate=true,width=2.700000in,height=2.700000in]{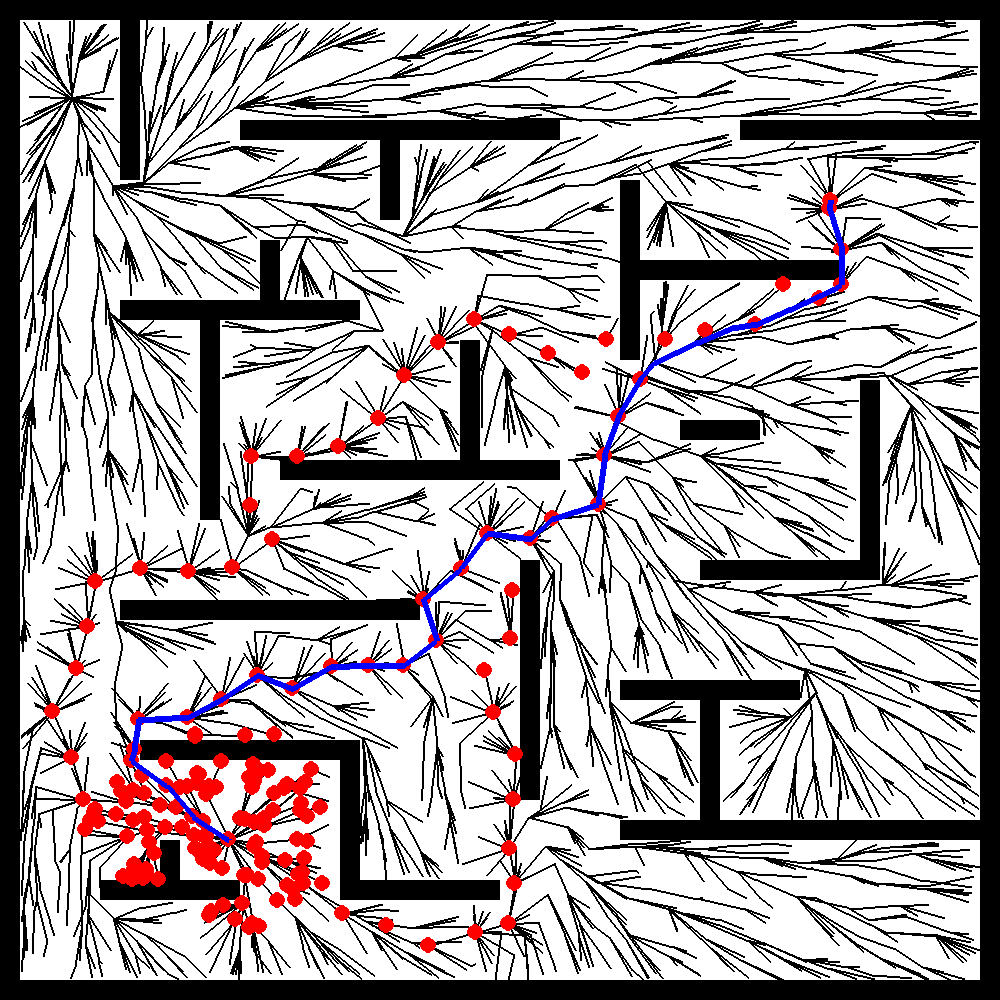}}%
\end{pgfscope}%
\begin{pgfscope}%
\pgfsetbuttcap%
\pgfsetmiterjoin%
\definecolor{currentfill}{rgb}{0.000000,0.741176,0.666667}%
\pgfsetfillcolor{currentfill}%
\pgfsetlinewidth{0.000000pt}%
\definecolor{currentstroke}{rgb}{0.000000,0.000000,0.000000}%
\pgfsetstrokecolor{currentstroke}%
\pgfsetstrokeopacity{0.000000}%
\pgfsetdash{}{0pt}%
\pgfpathmoveto{\pgfqpoint{9.959914in}{0.320934in}}%
\pgfpathlineto{\pgfqpoint{10.048341in}{0.320934in}}%
\pgfpathquadraticcurveto{\pgfqpoint{10.081675in}{0.320934in}}{\pgfqpoint{10.081675in}{0.354267in}}%
\pgfpathlineto{\pgfqpoint{10.081675in}{0.453033in}}%
\pgfpathquadraticcurveto{\pgfqpoint{10.081675in}{0.486366in}}{\pgfqpoint{10.048341in}{0.486366in}}%
\pgfpathlineto{\pgfqpoint{9.959914in}{0.486366in}}%
\pgfpathquadraticcurveto{\pgfqpoint{9.926581in}{0.486366in}}{\pgfqpoint{9.926581in}{0.453033in}}%
\pgfpathlineto{\pgfqpoint{9.926581in}{0.354267in}}%
\pgfpathquadraticcurveto{\pgfqpoint{9.926581in}{0.320934in}}{\pgfqpoint{9.959914in}{0.320934in}}%
\pgfpathclose%
\pgfusepath{fill}%
\end{pgfscope}%
\begin{pgfscope}%
\definecolor{textcolor}{rgb}{0.000000,0.000000,0.000000}%
\pgfsetstrokecolor{textcolor}%
\pgfsetfillcolor{textcolor}%
\pgftext[x=10.004128in,y=0.403650in,,]{\color{textcolor}\rmfamily\fontsize{8.000000}{9.600000}\selectfont A}%
\end{pgfscope}%
\begin{pgfscope}%
\pgfsetbuttcap%
\pgfsetmiterjoin%
\definecolor{currentfill}{rgb}{0.000000,0.741176,0.666667}%
\pgfsetfillcolor{currentfill}%
\pgfsetlinewidth{0.000000pt}%
\definecolor{currentstroke}{rgb}{0.000000,0.000000,0.000000}%
\pgfsetstrokecolor{currentstroke}%
\pgfsetstrokeopacity{0.000000}%
\pgfsetdash{}{0pt}%
\pgfpathmoveto{\pgfqpoint{11.523821in}{2.075934in}}%
\pgfpathlineto{\pgfqpoint{11.616434in}{2.075934in}}%
\pgfpathquadraticcurveto{\pgfqpoint{11.649768in}{2.075934in}}{\pgfqpoint{11.649768in}{2.109267in}}%
\pgfpathlineto{\pgfqpoint{11.649768in}{2.208033in}}%
\pgfpathquadraticcurveto{\pgfqpoint{11.649768in}{2.241366in}}{\pgfqpoint{11.616434in}{2.241366in}}%
\pgfpathlineto{\pgfqpoint{11.523821in}{2.241366in}}%
\pgfpathquadraticcurveto{\pgfqpoint{11.490488in}{2.241366in}}{\pgfqpoint{11.490488in}{2.208033in}}%
\pgfpathlineto{\pgfqpoint{11.490488in}{2.109267in}}%
\pgfpathquadraticcurveto{\pgfqpoint{11.490488in}{2.075934in}}{\pgfqpoint{11.523821in}{2.075934in}}%
\pgfpathclose%
\pgfusepath{fill}%
\end{pgfscope}%
\begin{pgfscope}%
\definecolor{textcolor}{rgb}{0.000000,0.000000,0.000000}%
\pgfsetstrokecolor{textcolor}%
\pgfsetfillcolor{textcolor}%
\pgftext[x=11.570128in,y=2.158650in,,]{\color{textcolor}\rmfamily\fontsize{8.000000}{9.600000}\selectfont G}%
\end{pgfscope}%
\end{pgfpicture}%
\makeatother%
\endgroup%
}
    
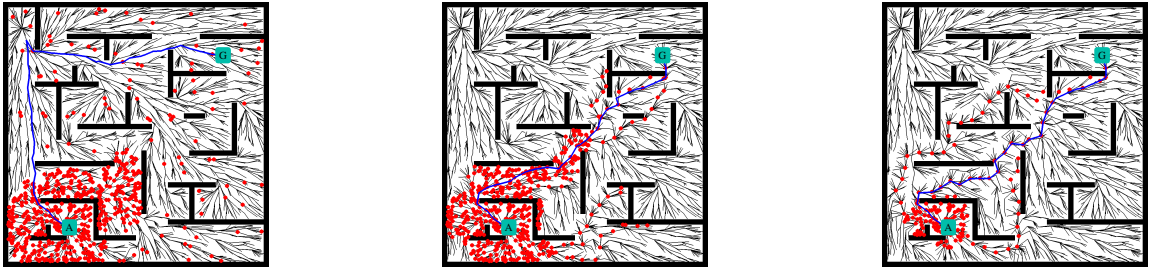
\captionof{figure}{Rewiring comparison using an identical tree between RT-RRT*, AM-RRT*(E) and AM-RRT*(D) (left, middle and right respectively), where (E) and (D) denote AM-RRT* using Euclidean and diffusion distance respectively as the assisting metric. After 750 nodes (shown in red) have been explored for rewiring, AM-RRT*(E) has significantly reduced the path cost in comparison to RT-RRT*. AM-RRT*(D) is able to achieve similar gains after exploring just 250 nodes. \\}
    \label{fig:figure}
\thispagestyle{empty}
\pagestyle{empty}
}]

\begin{abstract}

In this paper, we present a new algorithm that extends RRT* and RT-RRT* for online path planning in complex, dynamic environments. Sampling-based approaches often perform poorly in environments with narrow passages, a feature common to many indoor applications of mobile robots as well as computer games. Our method extends RRT-based sampling methods to enable the use of an assisting distance metric to improve performance in environments with obstacles. This assisting metric, which can be any metric that has better properties than the Euclidean metric when line of sight is blocked, is used in combination with the standard Euclidean metric in such a way that the algorithm can reap benefits from the assisting metric while maintaining the desirable properties of previous RRT variants - namely probabilistic completeness in tree coverage and asymptotic optimality in path length. We also introduce a new method of targeted rewiring, aimed at shortening search times and path lengths in tasks where the goal shifts repeatedly. We demonstrate that our method offers considerable improvements over existing multi-query planners such as RT-RRT* when using diffusion distance as an assisting metric: finding near-optimal paths with a decrease in search time of several orders of magnitude. Experimental results demonstrate our method offers a reduction of 99.5\% in planning times and 9.8\% in path lengths over RT-RRT* in a variety of environments. 
\end{abstract}

\section{INTRODUCTION}

\blfootnote{$^*$Research done during internship at Ocado Technology 10x}Multi-query, online planning in a mapped environment is a common task required for many motion planning applications from computer games \cite{games} to indoor delivery robots \cite{officerobot}. Such planning demands rapid search times as the goal or environment can change frequently and the agent must be able to respond without a large delay. Even in previously mapped environments, dynamic obstacles such as pedestrians or local changes such as the repositioning of furniture require new paths to be planned, highlighting the need for online planning. These real time planners must balance search time and path optimality, something that becomes increasingly difficult as the size and complexity of environment increases. Without an efficient planner there must be a choice made in large, complicated environments between delayed response time and poor quality paths.

Two of the most popular approaches to online planning are A* and RRT based methods \cite{critlit}. Algorithms involving potential fields \cite{potential} are also occasionally used for dynamic planning but are liable to become trapped in local minima. A* \cite{astar} has long been a staple in planning for computer games and real-time versions do exist \cite{realastar}, however it does require discretisation of the environment with the resolution significantly affecting search times. In practice, tactics such as computing navigation meshes \cite{navmesh} are employed to reduce the size of the search space for A*. RRT \cite{rrt} methods have also been adapted for real time planning, most notably CL-RRT \cite{clrrt} and RT-RRT* \cite{rtrrt}. The latter is the most promising RRT-based real-time planner to date, retaining the same tree between iterations (with the former only reusing a small subtree) to find paths without re-exploring areas already discovered. \blfootnote{\url{https://github.com/dan-armstrong/amrrt}}

In RRT-based planners, the Euclidean metric is traditionally used by itself but it can be a poor measure of pairwise potential between points without line of sight, which can lead the steering function to choose nearest neighbours that are unhelpful in expanding the graph (they may be on the other side of a wall, for example). Whilst other metrics are sometimes used instead of Euclidean (Palmieri and Arras propose a learned distance metric \cite{distancemetriclearning}) it is difficult to design a single metric that can both do as well as the Euclidean metric when there is line of sight (where the Euclidean metric is perfect) and also better when no line of sight exists. One contribution in this paper is enabling the use of a second assisting metric, used in conjunction with the Euclidean metric, that has better properties than the Euclidean metric when line of sight is blocked by obstacles. We choose diffusion distance \cite{difmapmotion} as the assisting metric to display the use of our algorithm as it offers a good approximation of the geodesic when line of sight is blocked - however it should be noted that our algorithm can be used with any assisting metric.

There are several contributions we make in this work. We first improve rewiring in RT-RRT* by introducing targeted rewiring that focuses on the nodes most likely to improve the path to the goal. We then add a preprocessing step that results in an assisting metric, before developing an extensions of both RRT* and RT-RRT* for use with such a metric whilst maintaining probabilistic completeness and same level of asymptotic path optimality found in RRT*/RT-RRT* regardless of the assisting metric chosen. The resulting algorithm is tested in various mapped environments to show empirically that it offers large speed-ups over existing planners and plans higher quality paths. We also produce a video demonstrating our method in action.\footnote{Link to video: \url{https://youtu.be/31yE8O9U6nQ}} Our method is designed to work well in large environments, a key limitation of RT-RRT* noted in Naderi, Rajam\"aki and H\"am\"al\"ainen's original paper \cite{rtrrt}. The tests also show that performance is improved by the rewiring improvements alone, without using an assisting metric.

\section{BACKGROUND}


\textbf{Rapidly-exploring Random Trees} (RRTs) \cite{rrt} are a class of sampling algorithms commonly used for single-query planning. They incrementally construct a tree representation of the environment by sampling points in the free space and connecting them along collision-free edges to a nearby node in the tree. The structure of the tree leads to trivial pathfinding back to the root, but regeneration of the tree is required whenever the agent moves to a new location - a costly process. Early versions were also not guaranteed to find optimal paths: something remedied with introduction of RRT* in 2011 \cite{rrtstar} which incorporated edge rewiring to ensure paths were asymptotically optimal. The more recent emergence of RT-RRT* \cite{rtrrt} offered the most promising real-time sampling planner to date. By retaining the same tree between queries and rewiring the root to move with the agent, near-optimal planning queries in dynamic environments could successfully be done online. However rewiring the entire tree becomes expensive as the tree grows larger, leading to suboptimal paths being found after the root changes location significantly. AM-RRT* remedies this by implementing targeted rewiring along the path to the goal.

\textbf{Diffusion maps} are a non-linear dimensionality reduction technique \cite{difmap}. One novel application of them is motion planning, proposed by Chen et al. in 2016 \cite{difmapmotion}. A grid graph of the environment is generated and used to construct a diffusion space that represents the underlying structure of the environment, with each state in the original environment mapped to a diffusion coordinate. The Euclidean distance between these diffusion coordinates (diffusion distance) is shown to approximate the pairwise potential between their respective states in the original environment. Path planning can then be done quickly via greedy methods. Whilst the diffusion map is successful in capturing the long range structures in the environment, small-scale details are lost when the dimensionality is reduced - leading to suboptimal paths. The more complex the environment, the more detail is lost and in practice this leads to consistently undesirable paths in large environments. By combining this information with rewiring techniques borrowed from sampling-based algorithms, AM-RRT* is able to utilize approximate information about the environment whilst retaining asymptotically optimal paths.

\section{PRELIMINARIES}

Let us denote the bounded space in which planning is done by $X \subseteq \mathbb{R}^d$, $d \in \mathbb{N}$. $X_{obs} \subset X$ is the set of all obstacles within the space, leaving $X_{free} = X \setminus X_{obs}$ as the free space. States in the environment are denoted by $\mathbf{x} \in X$ - for example the agent is represented by the state $\mathbf{x}_{agent}$ and the goal by $\mathbf{x}_{goal}$. The tree $T$ consists of nodes $\mathbf{x}_i \in X_{free}$. In addition, several user-defined constants are used to tune the algorithm. Together, $s_{max}$ (the maximum edge length) and $k_{max}$ (the maximum number of nodes within any circular neighbourhood of radius $s_{max}$) control the spread and density of $T$. The timing constants $t_{exp},\, t_{root}$, $t_{goal}$ and $t_{steer}$ are used to limit the amount of time spent expanding, root rewiring, goal rewiring and steering respectively - ensuring the algorithm remains real-time. The algorithm also requires an assisting metric $d_A$ on the space $X_{free}$ that aims to be an accurate metric to sort sets of points by so that they are approximately ordered by shortest path distance to the endpoint when the line of sight is blocked, as outlined in the introduction. Examples of suitable metrics are given in our approach. The metric $d_E$ is used to denote the standard Euclidean metric in $\mathbb{R}^d$. A-distance and E-distance are used as shorthand for distances in $d_A$ and $d_E$ respectively.

\section{APPROACH}

\subsection{Minor Algorithms}
Before we outline our method, the primitive procedures on which the main algorithms rely upon are described.

\textit{Sorting:} $Sort(X_s,\, \mathbf{x})$ returns the elements in $X_s$ sorted by order of increasing A-distance to $\mathbf{x}$. $Reverse(S)$ returns the sequence $S$ reversed. 

\textit{Stacks/queues:} $Enqueue(Q,\, \mathbf{x})$ adds $\mathbf{x}$ to the end of $Q$. $MultiEnqueue(Q,\, X)$ repeatedly calls $Enqueue(Q,\, \mathbf{x})$ for each $\mathbf{x}$ in the sequence $X$ (in sequence order). $Dequeue(Q)$ removes and returns the first item in $Q$. $Push(S,\, \mathbf{x})$ adds $\mathbf{x}$ to the front of $S$. $MultiPush(S,\, X)$ repeatedly calls $Push(S,\, \mathbf{x})$ for each $\mathbf{x}$ in the sequence $X$ (in sequence order). $Pop(S)$ removes and returns the first item in $S$. $Peek(S)$ returns but does not remove the first item in $S$. 

\textit{Trees:} $Tree(\mathbf{x})$ returns a tree with root node $\mathbf{x}$ and no edges. $AddNode(T,\, \mathbf{x}_{new},\, \mathbf{x}_{parent})$ adds the node $\mathbf{x}_{new}$ to $T$, connecting it to an existing node $\mathbf{x}_{parent}$ with an edge and returning the new tree. $Cost(T,\, \mathbf{x})$ returns the length of the path in $T$ from the root of the tree to $\mathbf{x}$, using the Euclidean metric. $Path(T,\, \mathbf{x})$ returns the sequence of nodes that form the path in $T$ from the root of the tree to $\mathbf{x}$. $SetRoot(T,\, \mathbf{x})$ returns the tree with the same nodes/edges as $T$ but rooted at $\mathbf{x}$, whilst $Root(T)$ returns $T$'s root node. $UpdateEdge(T,\, \mathbf{x}_n ,\, \mathbf{x}_c)$ replaces the edge $(\mathbf{x}_p,\, \mathbf{x}_c)$ with $(\mathbf{x}_n,\, \mathbf{x}_c)$ in $T$ and returns the new tree, where $\mathbf{x}_p$ is the parent of $\mathbf{x}_c$ in $T$.

\textit{Collision testing:} $ObsFree(\mathbf{x}_a, \mathbf{x}_b)$ returns true iff all of the points on the line between $\mathbf{x}_a$ and $\mathbf{x}_b$ lie in $X_{free}$.

\textit{Nearest neighbours:} $Nearest(T,\, \mathbf{x})$ returns the node in $T$ closest to $\mathbf{x}$. If the path between $\mathbf{x}$ and its Euclidean nearest neighbour $\mathbf{x}_{Enn}$ is free then $\mathbf{x}_{Enn}$ is returned, otherwise $\mathbf{x}$'s nearest neighbour as defined by $d_A$ is returned. $Nearby(T,\, \mathbf{x})$ returns the set of all nodes within E-distance $s_{max}$ of $\mathbf{x}$.

\textit{Geometry:} $Sphere(\mathbf{x},\, r)$ returns the set of states within the $d$-dimensional hypersphere (for $X \subseteq \mathbb{R}^d$) with center $\mathbf{x}$ and radius $r$. $RewireEllipse(T,\, \mathbf{x}_{goal})$ returns the set of states within the \textit{rewire ellipse} as defined in Informed RRT* \cite{informedrrt}: foci at $Root(T)$ and $\mathbf{x}_{goal}$, transverse diameter $c_{best}$ and conjugate diameters $\sqrt{c_{best}^2 - c_{min}^2}$, where $c_{best} = Cost(T,\, \mathbf{x}_{goal})$ and $c_{min} = d_E(Root(T),\, \mathbf{x}_{goal})$.

\textit{Sampling:} $Sample(X_s)$ returns an element sampled uniformly from the set $X_s$. $SampleState(T,\, \mathbf{x}_{goal})$ returns $Sample(X_s)$, with the set $X_{s}$ defined by the formula: 
\[   
X_{s} = 
     \begin{cases}
       \{\mathbf{x}_{goal}\} & p > \alpha \land \text{no path to } \mathbf{x}_{goal}\\
       X_{free} & p < \frac{\alpha}{\beta} \lor \text{no path to } \mathbf{x}_{goal}\\
       RewireEllipse(T,\, \mathbf{x}_{goal}) & \text{otherwise}\\
     \end{cases}
\]
$p \in [0,1]$ is a random number, with $\alpha \in [0,1]$ and $\beta \in [1, \infty)$ constants used to divide between the three methods of sampling. 

\subsection{Overview}
Our method extends RT-RRT* for use with an assisting metric as described in the introduction, leveraging the knowledge gained from it to explore the environment. However, our method can also be used for single-query planning by extending RRT* \cite{rrtstar} with the algorithms outlined in this section: choose $\mathbf{x}_{rand}$, $\mathbf{x}_{nrst}$ and $\mathbf{x}_{new}$ in RRT* using $SampleState$, $Nearest$ and $Steer$ as done in Algorithm 2. 

In this paper we use a metric based on diffusion distance $d_D$ as our assisting metric $d_A$. A diffusion map is generated as described in Chen et al. \cite{difmapmotion}: the environment is discretised as a grid graph from which a diffusion map is computed. Each state in the original grid graph is mapped to a diffusion state, with diffusion distance between two states in the original graph being the Euclidean distance between their respective diffusion states. From this we can define: $$d_D(\mathbf{x}_a,\, \mathbf{x}_b) = d_E(m(g(\mathbf{x}_a)),\, m(g(\mathbf{x}_b)))$$ where $g(\mathbf{x})$ maps a state $\mathbf{x}$ to its nearest grid point and $m(\mathbf{x})$ maps a state $\mathbf{x}$ in the grid graph to its diffusion coordinate. Diffusion distance offers a good approximation of pairwise potential with reasonable time and memory requirements (see results section for an empirical analysis).

\begin{algorithm}
\caption{AM-RRT*($\mathbf{x}_{agent},\, \mathbf{x}_{goal}$)}\label{Euclid}
\begin{algorithmic}[1]
\State $T \gets Tree(\mathbf{x}_{agent})$
\State $Q_{root} \gets [\, ]$; $Q_{goal} \gets [\, ]$; $S_{goal} \gets [\, ]$ 
\Loop
\State Update $\mathbf{x}_{agent},\, \mathbf{x}_{goal},\, X_{free},\, X_{obs}$
\While {time $< t_{exp}$}
\State $T \gets Expand(T,\, \mathbf{x}_{goal})$
\State $T \gets RewireRoot(T,\, Q_{root})$
\If {path to $\mathbf{x}_{goal}$ exists}
\State $T \gets RewireGoal(T,\, S_{goal},\, Q_{goal},\, \mathbf{x}_{goal})$
\EndIf
\EndWhile
\State $\mathbf{x}_{root},\, \mathbf{x}_1,\, \mathbf{x}_2$, ... $\gets Path(T,\, Nearest(T,\, \mathbf{x}_{goal}))$
\If {$d_E(\mathbf{x}_{agent},\, \mathbf{x}_{root}) < s_{max}$}
\State $T \gets SetRoot(T,\, \mathbf{x}_1)$
\EndIf
\State Move agent towards $\mathbf{x}_{root}$
\EndLoop
\end{algorithmic}
\end{algorithm}

AM-RRT* is outlined in Algorithm 1. The tree is initialised, rooted at the agent's position (line 1). By updating $\mathbf{x}_{goal}$ together with $X_{free}$ and $X_{obs}$ at the start of each iteration (line 4), we are able to accommodate multi-query planning (as the agent and goal can change) in dynamic environments (as the obstacle space can change). Whilst there is time to do so, the tree is expanded and rewired (lines 5-9) before a path is planned from the root to the goal (line 10). If there does not currently exist a path to the goal, the algorithm only rewires at the root. This allows the algorithm to dedicate more time to expansion when the agent either has no goal or has reached its destination, shifting between exploration and optimisation. Path planning can be done by finding the node closest to $\mathbf{x}_{goal}$ and tracing the path back to the root. If the agent is close to the root, the root is shifted along the path (lines 11-12). The agent can then be moved by any control algorithm (line 13). The time limit ensures the algorithm remains real-time, allowing the tree root to be updated and the agent moved at regular intervals.

\subsection{Tree Expansion}

Tree expansion in AM-RRT* follows a similar structure to previous RRT variants but differs in the way in which $\mathbf{x}_{rand}$ and $\mathbf{x}_{new}$ are generated. By interleaving nearest neighbour queries using both E-distance and A-distance, AM-RRT* is able to grow a tree that rapidly covers the space whilst still maintaining the desirable property of probabilistic completeness \cite{planningalgs}. This is ensured by defaulting to the Euclidean metric when possible: $\mathbf{x}_{nrst}$ is chosen to be the Euclidean nearest neighbour $\mathbf{x}_{E}$ if the line between $\mathbf{x}_{E}$ and $\mathbf{x}_{rand}$ is obstacle free, causing $\mathbf{x}_{new}$ to be projected along this line. This reduces tree expansion to the probabilistically complete expansion method seen in RRT* \cite{rrtstar} in these cases. 

\begin{algorithm}
\caption{Expand($T,\, \mathbf{x}_{goal}$)}\label{Euclid}
\begin{algorithmic}[1]
\State $\mathbf{x}_{rand} \gets SampleState(T,\, \mathbf{x}_{goal})$
\State $\mathbf{x}_{nrst} \gets Nearest(T,\, \mathbf{x}_{rand})$
\State $\mathbf{x}_{new} \gets Steer(\mathbf{x}_{nrst},\, \mathbf{x}_{rand})$
\State $X_{near} \gets Nearby(T,\, \mathbf{x}_{new})$
\If {$\vert X_{near} \vert \leq k_{max} \, \lor \, d_E(\mathbf{x}_{nrst},\, \mathbf{x}_{rand}) > s_{max}$}
\State $\mathbf{x}_{min} = \mathbf{x}_{nrst}$
\State $c_{min} = Cost(T,\, \mathbf{x}_{nrst}) + d_E(\mathbf{x}_{nrst},\, \mathbf{x}_{new})$
\For{$\mathbf{x}_{near}$ in $X_{near}$}
\State $c_{new} \gets Cost(T,\, \mathbf{x}_{near}) + d_E(\mathbf{x}_{near},\, \mathbf{x}_{new})$
\If {$c_{new} < c_{min} \land ObsFree(\mathbf{x}_{near},\, \mathbf{x}_{new})$}
\State $\mathbf{x}_{min} \gets \mathbf{x}_{near};\, c_{min} \gets c_{new}$
\EndIf
\EndFor
\State $T \gets AddNode(T,\, \mathbf{x}_{new},\, \mathbf{x}_{min})$
\EndIf
\State \Return $T$
\end{algorithmic}
\end{algorithm}

Algorithm 2 outlines tree expansion. A new state $\mathbf{x}_{rand}$ is randomly sampled (line 1) in a procedure similar to RT-RRT* \cite{rtrrt}, with $\mathbf{x}_{rand}$ chosen in one of three ways. If not yet reached, $\mathbf{x}_{goal}$ makes a good choice for $\mathbf{x}_{rand}$ as it ensures $\mathbf{x}_{new}$ is the point that best grows the tree towards the goal - allowing paths to be quickly constructed towards undiscovered goals. Sampling uniformly across the free space gives the tree good coverage of the entire environment, whereas targeted sampling from the rewire ellipse \cite{informedrrt} allows the current path to the goal to be improved when one exists. $\mathbf{x}_{new}$'s nearest neighbour $\mathbf{x}_{nrst}$ is then chosen by either E-distance or A-distance. Nearest neighbour queries in both Euclidean space and the assisting metric space can be carried out efficiently using dynamic vp-trees \cite{vptree} or other metric tree structures. To control the size of the tree, $\mathbf{x}_{new}$ only is added to the tree if its neighbourhood is not too full or if the E-distance between $\mathbf{x}_{nrst}$ and $\mathbf{x}_{new}$ is larger than $s_{max}$ (line 5). As in previous RRT algorithms, $\mathbf{x}_{new}$ is wired through the neighbour that minimises its cost-to-reach (lines 8-12).

\begin{algorithm}
\caption{Steer($\mathbf{x}_{init},\, \mathbf{x}_{end}$)}\label{Euclid}
\begin{algorithmic}[1]
\If {$ObsFree(\mathbf{x}_{init},\, \mathbf{x}_{end})$}
\State $p \gets 1 \,/\, max(1,\, d_E(\mathbf{x}_{init},\, \mathbf{x}_{end}))$
\State \Return $(1-p)\mathbf{x}_{init} + p\mathbf{x}_{end}$
\EndIf
\State $\mathbf{x}_{min} \gets \mathbf{x}_{init}$
\State $c_{min} \gets d_A(\mathbf{x}_{start},\, \mathbf{x}_{end})$
\State $X_s \gets Sphere(\mathbf{x}_{init},\, min(s_{max},\, d_E(\mathbf{x}_{init},\, \mathbf{x}_{end})))$
\While {time $< t_{steer}$}
\State $\mathbf{x}_{new} \gets Sample(X_s)$
\State $c_{new}  \gets d_A(\mathbf{x}_{new},\, \mathbf{x}_{end})$
\If {$c_{new} < c_{min} \, \land \, ObsFree(\mathbf{x}_{init}, \, \mathbf{x}_{new})$}
\State $\mathbf{x}_{min} \gets \mathbf{x}_{new};\, c_{min} \gets c_{new}$
\EndIf
\EndWhile
\State \Return $\mathbf{x}_{min}$
\end{algorithmic}
\end{algorithm}

Algorithm 3 describes how $\mathbf{x}_{new}$ is chosen to steer the tree towards $\mathbf{x}_{rand}$, returning a point that is closer (either by E-distance or A-distance) than $\mathbf{x}_{init}$ to $\mathbf{x}_{end}$. If the line from $\mathbf{x}_{init}$ to $\mathbf{x}_{end}$ is obstacle free, $\mathbf{x}_{new}$ is on along this line at maximum E-distance $s_{max}$ from $\mathbf{x}_{init}$ (lines 1-3). If the line is not obstacle free, random points are sampled (while time remains) at a maximum E-distance of $s_{max}$, or $d_E(\mathbf{x}_{init},\, \mathbf{x}_{end})$ if smaller, from $\mathbf{x}_{init}$ (lines 6-8). The point closest to $\mathbf{x}_{end}$ by A-distance that can be reached from $\mathbf{x}_{init}$ in free space is returned (lines 9 - 12). This second half of the algorithm leverages the information gained from the assisting metric to grow the tree in an informed manner in situations where previous RRT variants fail: when an obstacle blocks the path between the two states. The result is a method that is able to exploit the approximate information given by the metric whilst staying probabilistically complete. 

\subsection{Tree Rewiring}

Tree rewiring is used when the cost-to-reach of a node can be decreased by connecting it through a nearby node instead of its parent. It is also used to navigate around dynamic obstacles - if a node is blocked by an obstacle its cost (and subsequently the cost of its children) is set to $\infty$ which in turn causes these children to be rewired through neighbours with lower cost, creating an obstacle-free path to these nodes.

\begin{algorithm}
\caption{RewireRoot($T,\, Q_{root}$)}\label{Euclid}
\begin{algorithmic}[1]
\If {$Q_{root} \text{ not empty}$}
\State $Enqueue(Q_{root},\, Root(T))$
\EndIf
\While {time $< t_{root} \, \land \, Q_{root} \text{ not empty}$}
\State $\textbf{x}_{r} \gets Dequeue(Q_{root})$
\For {$\mathbf{x}_{near}$ in $Nearby(T,\, \mathbf{x}_{r})$}
\If {$ObsFree(\mathbf{x}_{near},\, \mathbf{x}_r)$}
\State $c_{old} \gets Cost(T,\, \mathbf{x}_{near})$
\State $c_{new} \gets Cost(T,\, \mathbf{x}_r) + d_E(\mathbf{x}_{r},\, \mathbf{x}_{near})$
\If {$c_{new} < c_{old}$}
\State $T \gets UpdateEdge(T,\, \mathbf{x}_r,\, \mathbf{x}_{near})$
\EndIf
\If {$\mathbf{x}_{near}$ not in $Q_{goal}$ since reset}
\State $Enqueue(Q_{goal},\, \mathbf{x}_{near})$
\EndIf
\EndIf
\EndFor
\EndWhile
\State \Return $T$
\end{algorithmic}
\end{algorithm}

Algorithm 4 is responsible for updating large portions of the tree when the root changes position. Once empty, the root queue is reset by adding $Root(T)$ (lines 1-2). The algorithm then rewires nodes out from the root whilst time remains and the queue is non-empty (line 3). $\mathbf{x}_r$ is dequeued and its neighbours $\mathbf{x}_{near}$ iterated through (lines 4-5). Each $\mathbf{x}_{near}$ is rewired through $\mathbf{x}_r$ if this reduces its cost (lines 7-10), with unseen nodes added to the queue (line 12). 

\begin{algorithm}
\caption{RewireGoal($T,\, S_{goal},\, Q_{goal},\,  \mathbf{x}_{goal}$)}\label{Euclid}
\begin{algorithmic}[1]
\If {$S_{goal} \text{ and } Q_{goal} \text{ empty}$}
\State $Push(S_{goal},\, Root(T))$
\EndIf
\While {time $< t_{goal} \, \land \, S_{goal} \text{ or } Q_{goal} \text{ not empty}$}
\State $\textbf{x}_r \gets \begin{cases}
       Pop(S_{goal}) & S_{goal} \text{ not empty}\\
       Dequeue(Q_{goal}) & Q_{goal} \text{ not empty}\\
     \end{cases}$
\If {$\mathbf{x}_r \in RewireEllipse(\mathbf{x}_{goal})$}
\State $X_{next} \gets \{\,\}$
\For{$\mathbf{x}_{near}$ in $Nearby(T,\, \mathbf{x}_r)$}
\If {$ObsFree(\mathbf{x}_{near},\, \mathbf{x}_r)$}
\State $c_{old} \gets Cost(T,\, \mathbf{x}_{near})$
\State $c_{new} \gets Cost(T,\, \mathbf{x}_r) + d_E(\mathbf{x}_{r},\, \mathbf{x}_{near})$
\If {$c_{new} < c_{old}$}
\State $T \gets UpdateEdge(T,\, \mathbf{x}_r,\, \mathbf{x}_{near})$
\EndIf
\If {$\mathbf{x}_{near}$ not in $S_{goal}$ since reset}
\State $X_{next} \gets X_{next} \cup \{\mathbf{x}_{near}\}$
\EndIf
\EndIf
\EndFor
\State $MultiPush(S_{goal},\, Reverse(Sort(X_{next},\, \mathbf{x}_{goal})))$
\State $MultiEnqueue(Q_{goal},\, Sort(X_{next},\, \mathbf{x}_{goal}))$
\EndIf
\If {$S_{goal}$ not empty}
\If {$d_A(Peek(S_{goal}),\mathbf{x}_{goal}) > d_A(\mathbf{x}_{r},\mathbf{x}_{goal})$}
\State $S_{goal} \gets [\, ]$
\EndIf
\EndIf
\EndWhile
\State \Return $T$
\end{algorithmic}
\end{algorithm}

Our method introduces a rewiring algorithm that specifically targets the path to the current goal (Algorithm 5). It uses both a stack $S_{goal}$ and queue $Q_{goal}$ of nodes to rewire along offshoot paths from the root to goal (two such offshoots are shown in Fig. 1). $S_{goal}$ stores nodes on the current offshoot with $Q_{goal}$ storing nodes from which the next offshoots will start. If both are empty, $S_{goal}$ is reset by pushing $Root(T)$. Only nodes within the rewire ellipse \cite{informedrrt} are considered (line 5) as only these can improve the current path to goal. Each node within radius $s_{max}$ of $\mathbf{x}_{r}$ is rewired if beneficial (lines 7 - 12), and then added to both $S_{goal}$ and $Q_{goal}$ if not yet explored (lines 15-16). These are sorted by A-distance to $\mathbf{x}_{goal}$ so that the first node retrieved from either is the one closest to $\mathbf{x}_{goal}$. If the next node in the offshoot (the node at the top of the stack) is further away than $x_r$, then the offshoot is discarded (lines 17-19).

By targeting nodes most likely to decrease path cost, AM-RRT* makes substantial decreases to path lengths in environments that are large or where the goal changes frequently. This is demonstrated in Fig. 1: rewiring is focused along offshoots towards the goal that dramatically shorten the path length to the goal, finding superior paths in half the time. This results in rapid shrinking of the rewire ellipse, meaning future rewiring efforts can be even more targeted. In RT-RRT* rewiring takes place sporadically across the graph, resulting in little improvement in the path to the goal.

\section{RESULTS}
\begin{figure*}
    \begin{center}
    \resizebox {\textwidth} {!} {
    \input{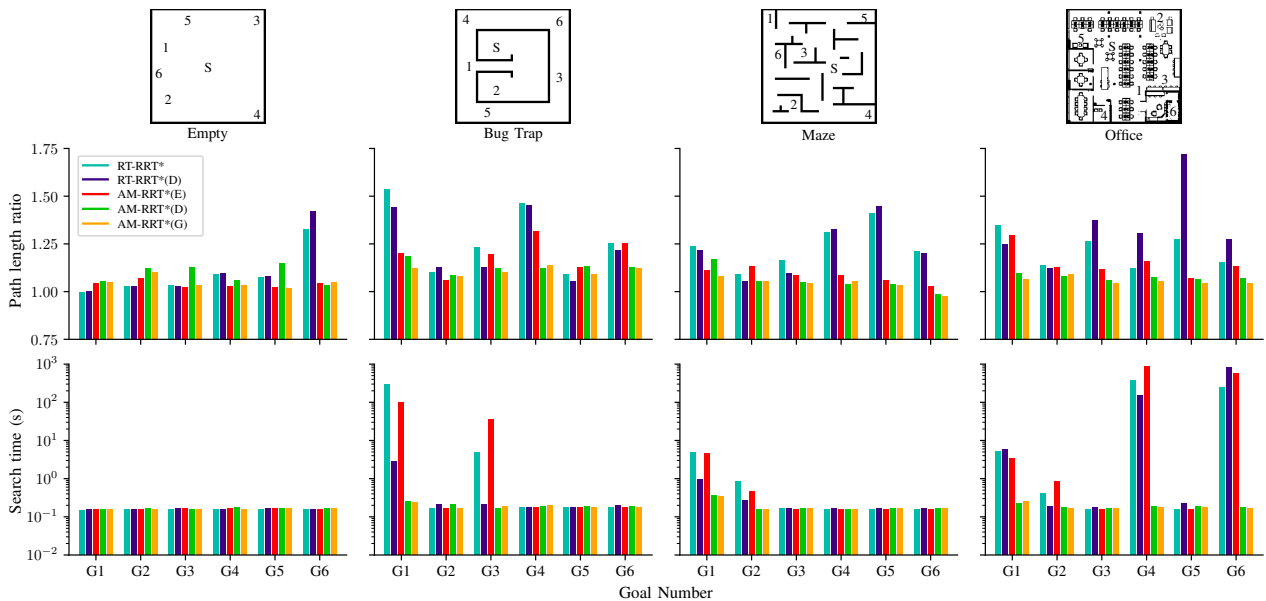}
    }
    \end{center}
    \caption{Performance statistics showing the distance of the path (relative to shortest path) taken between goals and the time taken (logarithmic scale) to find a path to the goal for the five planners in four different environments. On average, our chosen method AM-RRT*(D) is able to improve search times by 99.5\% and path lengths by 9.8\% over standard RT-RRT*. These improvements grow to 99.8\% and 11.7\% respectively when only considering the larger, complicated office environment.}
\end{figure*}

In the following section we compare our method against RT-RRT*. To achieve this we use five planners: RT-RRT*, RT-RRT*(D), AM-RRT*(E), AM-RRT*(D) and AM-RRT*(G). RT-RRT*(D) is RT-RRT* with diffusion distance used as the metric to find nearest neighbours instead of Euclidean distance. AM-RRT*(E), AM-RRT*(D) and AM-RRT*(G) are AM-RRT* planners using Euclidean, diffusion and geodesic distances respectively as the assisting metric. The tests are carried out in the four environments shown at the top of Fig. 2 (next page) with each experiment repeated 25 times and the results averaged. The environments are $100m \times 100m$, except Office which is $200m \times 200m$. For AM-RRT* planners we set $s_{max}$ as $5m$, $k_{max}$ as 20 and $t_{exp}$, $t_{root}$, $t_{goal}$ \& $t_{steer}$ as $0.15s$, $0.002s$, $0.004s$ \& $0.002s$ respectively. For RT-RRT* planners it was found in experiments that $k_{max}$ as 12 and $r_s$ as $5m$ gave the best results in the chosen environments. In order to keep the comparison consistent, in RT-RRT* the time for expansion and rewiring in each iteration was fixed at $0.15s$, with the times for root/random rewiring both fixed at $0.003s$. These values were found to give the best balance between exploration and rewiring. For fair measurement of the distances travelled, dynamic obstacles were not included in the testing environments.

We tested each planner on a typical task: start the agent at S and tour goals 1-6 in order, measuring the time taken to find a path to each goal and the length of the path from the previous goal. Our method offers substantial performance increases across all environments, as shown in Fig. 2. On average AM-RRT*(D) reduces paths lengths by 9.8\% and search times by 99.5\% over standard RT-RRT*, with these reductions becoming greater (11.7\% and 99.8\%, respectively) when only considering the office environment. The improvement in path quality can be explained by the rewiring updates made in our method: there is not enough time to update the entire tree between goals and subsequently the goal path is often not fully rewired to a near-optimal solution in RT-RRT* (see Fig. 1). 

The results also show the rewiring improvements boost planning performance without a better assisting metric, as AM-RRT*(E) reduces path lengths by 7.5\%. This does however come with an increase in planning times by a factor of 1.68, which can be explained by the fixed max edge included in AM-RRT* - in practice it has been found that this greatly improves rewiring capabilities whilst slightly reducing exploration speed. This is a worthwhile compromise, as the planner is designed for extended use in one environment.  AM-RRT* also adds new nodes at a slower rate than RT-RRT*, spending more time choosing how to connect new nodes to the graph. This is why AM-RRT* performs slightly worse than RT-RRT* when finding paths to goals (1) and (2) in the empty environment: RT-RRT* samples nodes more quickly and is able to build a smoother path. This is only an issue that becomes apparent in the most trivial cases - the results show on the whole it is beneficial to add new nodes better, not faster.

The benefits of interleaving the two metrics done in our method can be seen through the performance of RT-RRT*(D). By naively integrating the diffusion map without including the modifications made to steering and rewiring in our method, RT-RRT*(D) depends entirely on the assisting metric for exploration - even in situations where this metric may be unreliable. We can see this in the office environment: the diffusion map is less accurate due to the complex nature of the environment, leaving exploration times several orders of magnitude (868 times) larger than AM-RRT*(D) despite them both using the same diffusion map. Without the modifications made in AM-RRT*, probabilistic completeness is not guaranteed and RT-RRT*(D) is liable to become trapped.\footnote{This occurred in a small number of tests in the office environment for RT-RRT*(D), these cases were left out of the results thus making the actual performance of RT-RRT*(D) worse than what is shown.}

An important value not included in the results is the amount of preprocessing required for each of the metrics. It took an average of just 26.1s to generate each diffusion map, with 2723s needed to generate each matrix of geodesic pairwise potentials for the same grid of nodes. The Euclidean metric required no preprocessing. When including the time required for preprocessing, AM-RRT*(D) is still 31 times faster than RT-RRT* - making it an attractive option as a planner operating in new enviroments as well as one repeatedly operating in the same environment. The time required to generate the distance matrix for the geodesic matrix means it is rarely a useful assisting metric in practice: AM-RRT*(G) is only included in the results to show an upper bound of our method. \footnote{The nature of random sampling means this upper bound is not perfect: in the empty environment when finding a path to (6), AM-RRT*(G) finds marginally longer paths than AM-RRT*(D).} Fig. 2 makes it clear to see that by using diffusion maps we can get close-to-perfect results (path lengths using AM-RRT*(D) are just 8.9\% longer than optimal) with minimal preprocessing. 

\section{CONCLUSION}
In this paper, we proposed extensions of RRT* and RT-RRT* that use an assisting metric to better explore large, complex environments. Probabilistic completeness and asymptotic optimality were achieved by combining the assisting metric with the standard Euclidean metric. We also introduced a new rewiring technique that targets the goal path, rewiring along offshoots from the agent out to the goal. This rewiring is further concentrated by only considering nodes that have the potential to improve the current goal path. We also introduced a new steering algorithm to best grow the tree, interleaving information from both the Euclidean and assisting metric. By performing tests in a range of environments we showed empirically that our method offers large performance gains, most significantly in the complex, narrow environments likely to be encountered in real life applications of indoor robots and  computer game agents. 

AM-RRT* does have several limitations, the largest of which being the fact many useful assisting metrics, including diffusion distance, require the environment to be mapped in advance as the metric will need to have an understanding of the environment's geometry to yield large increases in planning speed. This makes it less suitable for use in unexplored environments (although it will still work here using the Euclidean metric as the assisting metric). The assisting metric may also require a substantial amount of preprocessing, meaning it is more applicable as a planner for agents that operate in the same area(s) repeatedly, such as domestic/factory robots or characters in a computer game. 

A potential avenue for research is to speed up the preprocessing step by leveraging learning based approaches such as deep learning, possibly by training a neural network that takes the map as input and outputs a grid based map (like a diffusion map) where the distance between each point approximates the shortest path distance between each point in the original map, or is encouraged to be a good distance to sort sets of points by to have them approximately ordered by shortest path distance to the endpoint.

\printbibliography

@article{astar,
  title={A formal basis for the heuristic determination of minimum cost paths},
  author={Hart, Peter E and Nilsson, Nils J and Raphael, Bertram},
  journal={IEEE transactions on Systems Science and Cybernetics},
  volume={4},
  number={2},
  pages={100--107},
  year={1968},
  publisher={IEEE}
}

@article{rrt,
  title={Rapidly-exploring random trees: A new tool for path planning},
  author={LaValle, Steven M},
  year={1998},
  publisher={Citeseer}
}

@article{rrtstar,
  title={Sampling-based algorithms for optimal motion planning},
  author={Karaman, Sertac and Frazzoli, Emilio},
  journal={The international journal of robotics research},
  volume={30},
  number={7},
  pages={846--894},
  year={2011},
  publisher={Sage Publications Sage UK: London, England}
}

@inproceedings{rtrrt,
  title={RT-RRT*: A real-time path planning algorithm based on RRT},
  author={Naderi, Kourosh and Rajam{\"a}ki, Joose and H{\"a}m{\"a}l{\"a}inen, Perttu},
  booktitle={Proceedings of the 8th ACM SIGGRAPH Conference on Motion in Games},
  pages={113--118},
  year={2015},
  organization={ACM}
}

@inproceedings{difmapmotion,
  title={Motion planning with diffusion maps},
  author={Chen, Yu Fan and Liu, Shih-Yuan and Liu, Miao and Miller, Justin and How, Jonathan P},
  booktitle={2016 IEEE/RSJ International Conference on Intelligent Robots and Systems (IROS)},
  pages={1423--1430},
  year={2016},
  organization={IEEE}
}

@inproceedings{informedrrt,
  title={Informed RRT*: Optimal sampling-based path planning focused via direct sampling of an admissible ellipsoidal heuristic},
  author={Gammell, Jonathan D and Srinivasa, Siddhartha S and Barfoot, Timothy D},
  booktitle={2014 IEEE/RSJ International Conference on Intelligent Robots and Systems},
  pages={2997--3004},
  year={2014},
  organization={IEEE}
}

@article{difmap,
  title={Diffusion maps},
  author={Coifman, Ronald R and Lafon, St{\'e}phane},
  journal={Applied and computational harmonic analysis},
  volume={21},
  number={1},
  pages={5--30},
  year={2006},
  publisher={Elsevier}
}

@inbook{planningalgs,
  title={Planning algorithms},
  author={LaValle, Steven M},
  year={2006},
  publisher={Cambridge university press},
 pages = {185-186}
}

@article{vptree,
  title={Dynamic vp-tree indexing for n-nearest neighbor search given pair-wise distances},
  author={Fu, Ada Wai-chee and Chan, Polly Mei-shuen and Cheung, Yin-Ling and Moon, Yiu Sang},
  journal={The VLDB Journal},
  volume={9},
  number={2},
  pages={154--173},
  year={2000},
  publisher={Springer}
}

@article{games,
  title={High quality navigation in computer games},
  author={Nieuwenhuisen, Dennis and Kamphuis, Arno and Overmars, Mark H},
  journal={Science of Computer Programming},
  volume={67},
  number={1},
  pages={91--104},
  year={2007},
  publisher={Elsevier}
}

@article{officerobot,
  title={DERVISH an office-navigating robot},
  author={Nourbakhsh, Illah and Powers, Rob and Birchfield, Stan},
  journal={AI magazine},
  volume={16},
  number={2},
  pages={53--53},
  year={1995}
}

@article{potential,
  title={Dynamic motion planning for mobile robots using potential field method},
  author={Ge, Shuzhi Sam and Cui, Yun J},
  journal={Autonomous robots},
  volume={13},
  number={3},
  pages={207--222},
  year={2002},
  publisher={Springer}
}

@article{critlit,
  title={A critical evaluation of literature on robot path planning in dynamic environment},
  author={Rastgoo, Mohammad Naim and Nakisa, Bahareh and Nasrudin, Mohammad Faidzul and Nazri, Mohd Zakree Ahmad},
  journal={Journal of Theoretical and Applied Information Technology},
  volume={70},
  number={1},
  pages={177--185},
  year={2014},
  publisher={Asian Research Publishing Network (ARPN)}
}

@article{realastar,
  title={Real-time heuristic search for motion planning with dynamic obstacles},
  author={Cannon, Jarad and Rose, Kevin and Ruml, Wheeler},
  journal={Ai Communications},
  volume={27},
  number={4},
  pages={345--362},
  year={2014},
  publisher={IOS Press}
}

@inproceedings{clrrt,
  title={Bounds on tracking error using closed-loop rapidly-exploring random trees},
  author={Luders, Brandon D and Karaman, Sertac and Frazzoli, Emilio and How, Jonathan P},
  booktitle={Proceedings of the 2010 American Control Conference},
  pages={5406--5412},
  year={2010},
  organization={IEEE}
}

@article{navmesh,
  title={A navigation mesh for dynamic environments},
  author={Van Toll, Wouter G and Cook IV, Atlas F and Geraerts, Roland},
  journal={Computer Animation and Virtual Worlds},
  volume={23},
  number={6},
  pages={535--546},
  year={2012},
  publisher={Wiley Online Library}
}

@inproceedings{distancemetriclearning,
  title={Distance metric learning for RRT-based motion planning for wheeled mobile robots},
  author={Palmieri, Luigi and Arras, Kai O},
  booktitle={2014 IROS Machine Learning in Planning and Control of Robot Motion Workshop},
  year={2014}
}

\addtolength{\textheight}{-12cm}   





\end{document}